\pdfoutput=1

\documentclass[11pt]{article}

\usepackage[final]{acl}

\usepackage{times}
\usepackage{latexsym}
\usepackage{amsmath}

\usepackage[T1]{fontenc}

\usepackage[utf8]{inputenc}

\usepackage{microtype}

\usepackage{inconsolata}

\usepackage{graphicx}

\usepackage{booktabs}
\usepackage{multirow}
\usepackage{graphicx}
\usepackage{xcolor}
\usepackage[normalem]{ulem}
\useunder{\uline}{\ul}{}

\usepackage{pifont}
\usepackage{xspace}
\usepackage{amssymb}

\definecolor{addgreen}{HTML}{007A33}   
\definecolor{delred}{HTML}{ff3100}     
\definecolor{reporange}{HTML}{B35900}  
\definecolor{appblue}{HTML}{000AFF}  

\newcommand{\cmark}[1]{%
  {\textcolor[HTML]{#1}{\ding{51}}}\xspace%
}

\newcommand{\xmark}[1]{%
  {\textcolor[HTML]{#1}{\ding{55}}}\xspace%
}

\title{LLMs cannot spot math errors, even when allowed to peek into the solution}

\author{
    KV Aditya Srivatsa\\
    \And
    Kaushal Kumar Maurya\\
    Mohamed bin Zayed University of Artificial Intelligence, Abu Dhabi, UAE \\
    {\tt kvaditya.edu@gmail.com,} {\tt \{kaushal.maurya, ekaterina.kochmar\}@mbzuai.ac.ae} \\
    \And
    Ekaterina Kochmar
}

\begin{document}
\maketitle
\begin{abstract}
Large language models (LLMs) demonstrate remarkable performance on math word problems, yet they have been shown to struggle with \textit{meta-reasoning} tasks such as identifying errors in student solutions. In this work, we investigate the challenge of locating the first error step in stepwise solutions using two error reasoning datasets: {\tt VtG} and {\tt PRM800K}. Our experiments show that state-of-the-art LLMs struggle to locate the first error step in student solutions even when given access
to the reference solution. To that end, we propose an approach that generates an intermediate \textit{corrected student solution}, aligning more closely with the original student's solution, which helps improve performance.\footnote{All data and code are available at \url{https://github.com/kvadityasrivatsa/llms-cannot-spot-math-errors}}

\end{abstract}

\section{Introduction}
\label{sec:introduction}

Large language models (LLMs) demonstrate impressive performance on existing reasoning benchmarks, particularly on math word problems \cite{liu2024deepseek, dubey2024llama, achiam2023gpt}. For example, the state-of-the-art {\tt Llama3.1-405B} \cite{dubey2024llama}  model achieves 96.8\% accuracy on the challenging {\tt GSM8K} reasoning benchmark \cite{cobbe2021trainingverifierssolvemath}. However, recent work has revealed that models excelling at end-task accuracy often fail when probed about their underlying reasoning processes -- what we refer to here as {\em meta-reasoning}. For instance, both \citet{zeng2024mrgsm8kmetareasoningbenchmarklarge} and \citet{tyen-etal-2024-llms} have reframed LLMs from passive problem solvers into active evaluators, revealing that even top-performing LLMs struggle with tasks like locating the first error step in a student's solution.

\begin{figure}[t]
     \centering
     \includegraphics[width=0.95\columnwidth]{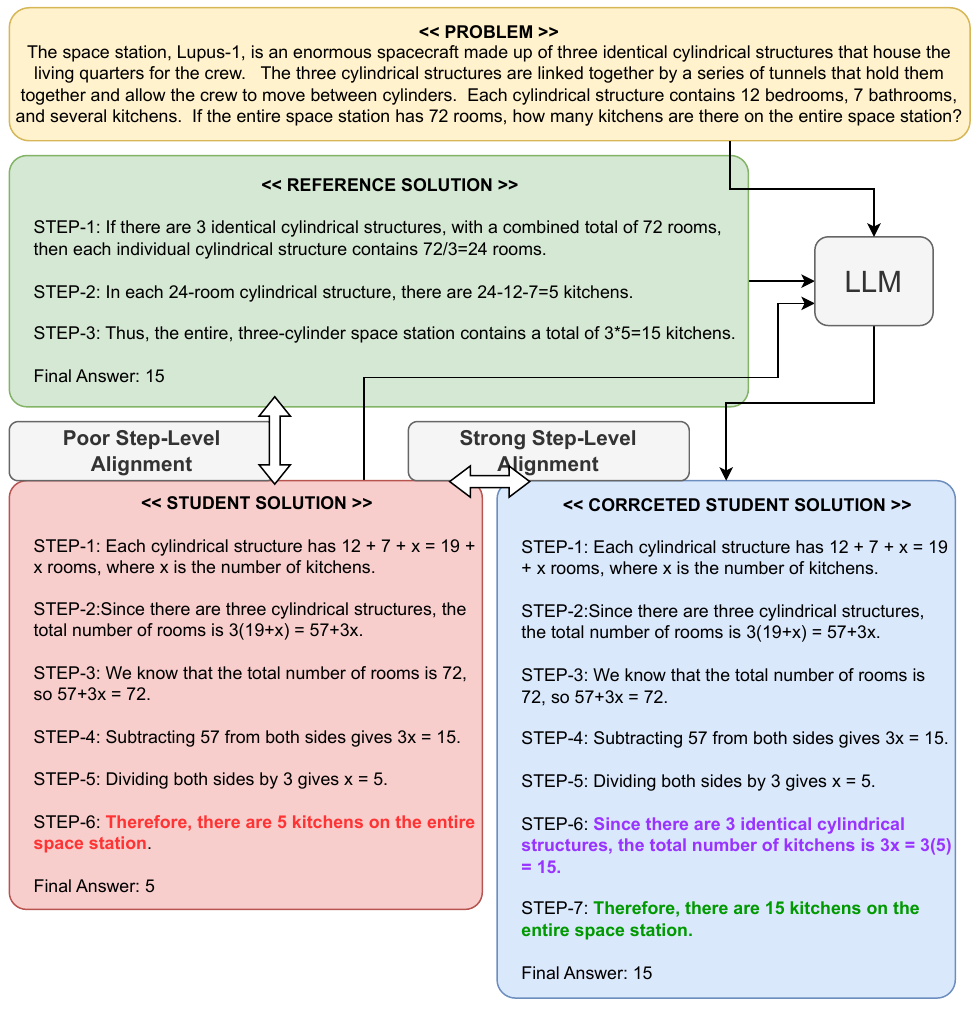}
     \caption{In this example, the \textit{corrected} version of the original \textit{student solution} (with {\tt Llama3-70B}) makes the location of the student's first mistake more apparent as compared to the reference (i.e., gold) solution.}
     \label{fig:correction_over_gs}
     \vspace{-10pt}
\end{figure}

The ability to pinpoint and categorize errors is not only a critical diagnostic tool for understanding models' limitations but is also essential for developing assistive educational feedback tools in intelligent tutoring systems \cite{jia2024assessing, nino2024systematic}. Accurate error detection and categorization promote better personalized and effective feedback generation \cite{anderson1990cognitive, hattie2007power}. This was demonstrated using LLMs by \citet{daheim-etal-2024-stepwise}, where decomposing the feedback generation process to verify the student's solution before providing hints produces more reliable results. Existing research typically tests LLMs on \textit{error localization} using only the original question and the student's erroneous solution \cite{zeng2024mrgsm8kmetareasoningbenchmarklarge, tyen-etal-2024-llms}. Pedagogical research indicates that when teachers have a canonical solution, they can “offload” the problem-solving process and focus on comparing student work against the expert path \cite{sweller1998cognitive, carpenter1989cognition}. Motivated by this, we investigate \textbf{what information helps an LLM in locating the first error step in a math problem solution}. Our preliminary experiments suggest that even when gold (reference) solution is provided, LLMs still struggle.

Therefore, we explore an alternative approach that generates an intermediate corrected version of the student's solution. This version preserves the student's method while applying only necessary structural adjustments, yielding a reference that both mirrors the student's reasoning and maintains consistency. We call it \textit{corrected student solution}. Figure \ref{fig:correction_over_gs} shows a math problem from {\tt GSM8K} with its reference solution, an incorrect stepwise student solution from the {\tt VtG} \cite{daheim-etal-2024-stepwise} dataset, and a corrected version produced by {\tt Llama3-70B} LLM \cite{grattafiori2024llama3herdmodels}. Aligning the student’s solution with the ground truth is crucial to pinpoint the first error but is challenging due to: \textbf{(1) Poor Step Alignment}: the student’s 6 steps versus the reference's 3 steps do not correspond directly, and there are no matching intermediate variables until the 4th step; and \textbf{(2) Different Approaches}: the student introduces an unknown variable \textit{x}, while the reference follows a more direct method. The corrected student solution shows that although the student computes the number of kitchens per cylindrical structure correctly, they overlook calculating the total number of kitchens in the space station. By updating the solution to mirror the reference while retaining the student's approach, the corrected solution achieves better step alignment, simplifying error identification and error localization. Our analysis (see \S\ref{subsec:app-bertscore-alignment}) shows that these generated corrected student solutions semantically align with student solutions better than the gold solutions.

In this paper, we formulate and investigate two key research questions: {\bf RQ1}: Can LLMs accurately locate errors in incorrect math problem solutions when provided with access to the reference solution? and {\bf RQ2}: Can the incorporation of intermediate reasoning steps -- such as corrected student solution -- enhance the overall performance of LLMs in the task of error localization?

Our experiments on two public datasets -- {\tt VtG} \cite{daheim-etal-2024-stepwise} and {\tt PRM800K} \cite{lightman2023letsverifystepstep} -- confirm that state-of-the-art models like {\tt Llama3.1-405B} and {\tt GPT-4o} face significant difficulties in accurately localizing the first error even when furnished with the dataset-provided gold solution. In contrast, supplying a corrected student solution markedly improves error localization performance, especially for more capable models, which suggests that overall problem solving ability has little bearing on error detection accuracy.

\section{Methodology}
\label{sec:method}

\paragraph{Data} We perform our experiments on two error-reasoning datasets to investigate LLMs’ capabilities across varying levels of problem difficulty and error typologies. The first dataset, referred to in this work as {\tt VtG}, was released by \citet{daheim-etal-2024-stepwise} and comprises 1,002 incorrect stepwise student attempts on grade school-level math word problems in English. These attempts are sourced from {\tt MathDial} \cite{macina-etal-2023-mathdial} and originally from {\tt GSM8K} \cite{cobbe2021trainingverifierssolvemath}, and include annotations for the first erroneous step, a description of the mistake, and its classification into one of seven error types (see §\ref{sec:app-data-details} for more details). The second dataset, {\tt PRM800K} \cite{lightman2023letsverifystepstep}, consists of 80,000 incorrect stepwise student solutions -- with 2,077 designated for testing -- each marked at the first error step, and features math questions drawn from the {\tt MATH} \cite{hendrycks2021measuringmathematicalproblemsolving} dataset, with more advanced problems than those in {\tt GSM8K}. Together, these datasets present diversity to explore error reasoning and meta-reasoning across elementary and advanced math levels.

\paragraph{LLMs} We select a diverse array of 6 open and closed-source, as well as generic and fine-tuned LLMs for our experiments (listed in Table \ref{tab:llm-listing}). Notably, {\tt Qwen2.5-72B-Math} \cite{yang2024qwen25mathtechnicalreportmathematical} has been fine-tuned for solving math problems, and {\tt LearnLM-1.5-Pro} \cite{learnlmteam2024learnlmimprovinggeminilearning} has been built for advanced pedagogical reasoning, guiding mistake discovery and providing constructive feedback. In addition to these properties, these models were chosen based on their problem-solving performance on the underlying math problems in our two datasets (see definition and scores in §\ref{subsec:app-problem-solving-performance}). In particular, while most models in our selection excel at the grade-school arithmetic problems from {\tt GSM8K} \cite{cobbe2021trainingverifierssolvemath}, they exhibit varied performance on the more advanced questions from the {\tt MATH} \cite{hendrycks2021measuringmathematicalproblemsolving} dataset.

\paragraph{Modeling Approach}
Let \(Q\) be the problem, \(G\) the reference solution, \(S=\{s_i\}_{i=1}^{n}\) the student trace, and \(E\) the first erroneous step.  
An LLM\(_{\theta}\) with prompt template \(P\) predicts \(E\) under three settings:  
(i) \textbf{problem + student} — \(E=\mathrm{LLM}_{\theta}\bigl(P(Q,S)\bigr)\);  
(ii) \textbf{problem + student + gold} — \(E=\mathrm{LLM}_{\theta}\bigl(P(Q,S,G)\bigr)\);  
(iii) \textbf{problem + student + correction} — first align the gold to the student trace,
\(S'=\{s'_j\}_{j=1}^{m}=\mathrm{LLM}_{\theta}\bigl(P(G,S)\bigr)\), producing a structurally and stylistically matched correction of \(S\)  (see~\S\ref{subsec:app-bertscore-alignment}); then detect the error with \(Q,S,S'\):  
\(E=\mathrm{LLM}_{\theta}\bigl(P(Q,S,S')\bigr)\).

\paragraph{Gold Solution vs. Corrected Solution}
Among recent work, the approach of \citet{li2024askbeforedetectionidentifyingmitigatingconformity} is most similar to ours. They ask the model to generate a corrected version of the student’s solution from scratch and then localize errors against that self-generated reference, so they find that their success depends on the model’s own problem-solving ability. Instead, we cast the model as a “teacher”: it is given the gold solution and tasked with generating a corrected version of the student's solution. Providing the gold answer disentangles error detection abilities from problem solving abilities. Since gold solutions in benchmarks like {\tt GSM8K} and {\tt MATH} often differ in style, step order, and content, we include a brief intermediate step to rewrite the gold solution to closely match that of the student, making only minimal edits needed for correctness.

To evaluate the quality of these corrections (\(S'\)), we manually annotated 90 randomly selected outputs across models for (1) correctness (overall and step-level) and (2) stylistic similarity to the student’s work. A subset of 30 was double-annotated (Cohen’s $\kappa$ = 0.82 and 0.85 for correctness and stylistic similarity) (see §\ref{subsec:app-manual-verification} for more details). Most models produced accurate corrections in over 93.3\% of cases and maintained stylistic similarity in 87.4\%. The exception was {\tt Qwen2.5-72B-Math}, which scored significantly lower on both metrics (69.6\% correctness, 63.3\% stylistic similarity), consistent with its weaker error localization performance (see §\ref{subsec:exact-error-step-performance}).

\section{Experimental Setup}
To generate first error step predictions, we adapt the few-shot prompt from \citet{daheim-etal-2024-stepwise} and define four prompt types, as described in Section \ref{sec:method}: \textbf{(1) \textit{w/o-S}} (without gold solution) presents only the math problem and the student's incorrect stepwise solution, asking the LLM to identify the first error; \textbf{(2) \textit{w-GS}} (with gold solution) additionally provides the dataset’s stepwise gold solution; \textbf{(3) \textit{w-Cor}} (with corrected student solution) first prompts the LLM to generate a corrected version of the student’s solution—retaining their approach but fixing errors using the problem and gold solution—and then uses this in the main prompt; and \textbf{(4) random} selects a random error step within the student's solution span, averaged over 100 runs with different seeds. We also evaluate each LLM’s problem-solving ability on both datasets to compare against their error localization performance. Exact prompt templates, LLM settings, and other details are provided in §\ref{subsec:app-prompts}.

\begin{table}[]
\centering
\resizebox{\columnwidth}{!}{%
\begin{tabular}{@{}lccc|ccc@{}}
\toprule
\textbf{Model} & \multicolumn{3}{c|}{\textbf{{\tt VtG}}} & \multicolumn{3}{c}{\textbf{{\tt PRM800K}}} \\ \midrule
{\tt Random} & \multicolumn{3}{c|}{18.32} & \multicolumn{3}{c}{9.52} \\ \midrule
 & \textit{w/o-S} & \textit{w-GS} & \textit{w-Cor} & \textit{w/o-S} & \textit{w-GS} & \textit{w-Cor} \\ \cmidrule(l){2-7} 
{\tt Llama3-70B} & 42.51 & 49.50 & \textbf{61.28} & 19.64 & 24.12 & \textbf{33.03} \\
{\tt Llama3.1-70B} & 49.10 & 57.98 & \textbf{64.17} & 24.46 & 34.23 & \textbf{38.39} \\
{\tt Llama3.1-405B} & 49.90 & 62.38 & {\ul \textbf{64.77}} & 24.12 & 39.54 & \textbf{47.86} \\
{\tt GPT-4o} & 54.49 & 63.57 & \textbf{64.57} & 39.29 & 43.72 & \textbf{49.40} \\ 
{\tt Qwen2.5-72B-Math} & {\bf 45.01} & 30.44 & 19.10 & 21.86 & {\bf 28.50} & 21.47 \\ 
{\tt LearnLM-1.5-Pro} & 54.89 & {\bf 64.07} & 63.67 & 42.51 & 49.69 & {\ul \textbf{51.13}} \\ 
\bottomrule
\end{tabular}%
}
\vspace{-0.2cm}
\caption{\small First error step localization accuracy (in \%) on {\tt VtG} and {\tt PRM800K} datasets. For each task, within each dataset, the {\bf bold} value represents the highest accuracy per LLM, whereas the {\ul underlined} value represents the overall highest accuracy.}
\vspace{-0.4cm}
\label{tab:error-diagnosis-results}
\end{table}

\section{Results and Analyses}

\subsection{Exact Error Step Prediction}
\label{subsec:exact-error-step-performance}

Table \ref{tab:error-diagnosis-results} shows error step prediction accuracies for all model and prompt type combinations across both datasets. In general, scores are higher for {\tt VtG} than {\tt PRM800K}—likely due to 
a greater number of steps per solution on average in {\tt PRM800K} (13.3) than in {\tt GSM8K} (5.9). Score variation is larger in {\tt PRM800K}, with smaller models like {\tt Llama3-70B} and {\tt Llama3.1-70B} performing comparably to larger ones on {\tt VtG}. Without any reference solution (\textit{w/o-S}), accuracy remains low, as reported in previous studies \cite{zeng2024mrgsm8kmetareasoningbenchmarklarge, tyen-etal-2024-llms}. \textit{Although providing the gold solution (w-GS) increases accuracy, most models still struggle to pinpoint the exact error step} (see \textbf{RQ1}). The \textit{corrected solution (w-Cor) improves performance over w-GS} and yields the highest accuracy across most models for both datasets (see \textbf{RQ2}). Interestingly, {\tt LearnLM-1.5-Pro} shows almost no gain from intermediate corrections: its \textit{w-GS} accuracy slightly surpasses \textit{w-Cor} on {\tt VtG}. This likely reflects the model’s prior fine-tuning for mistake detection and feedback generation, which already leverages the gold solution signal, leaving little headroom for additional corrections. In sharp contrast, {\tt Qwen2.5-72B-Math}—tuned for problem solving rather than critique—records the lowest accuracies overall and even drops in both \textit{w-GS} and \textit{w-Cor} compared to \textit{w/o-S}, while scoring highly in problem-solving on both datasets (see Table \ref{tab:solving_acc}). 

\begin{figure*}[t!]
     \centering
    \includegraphics[width=\textwidth]{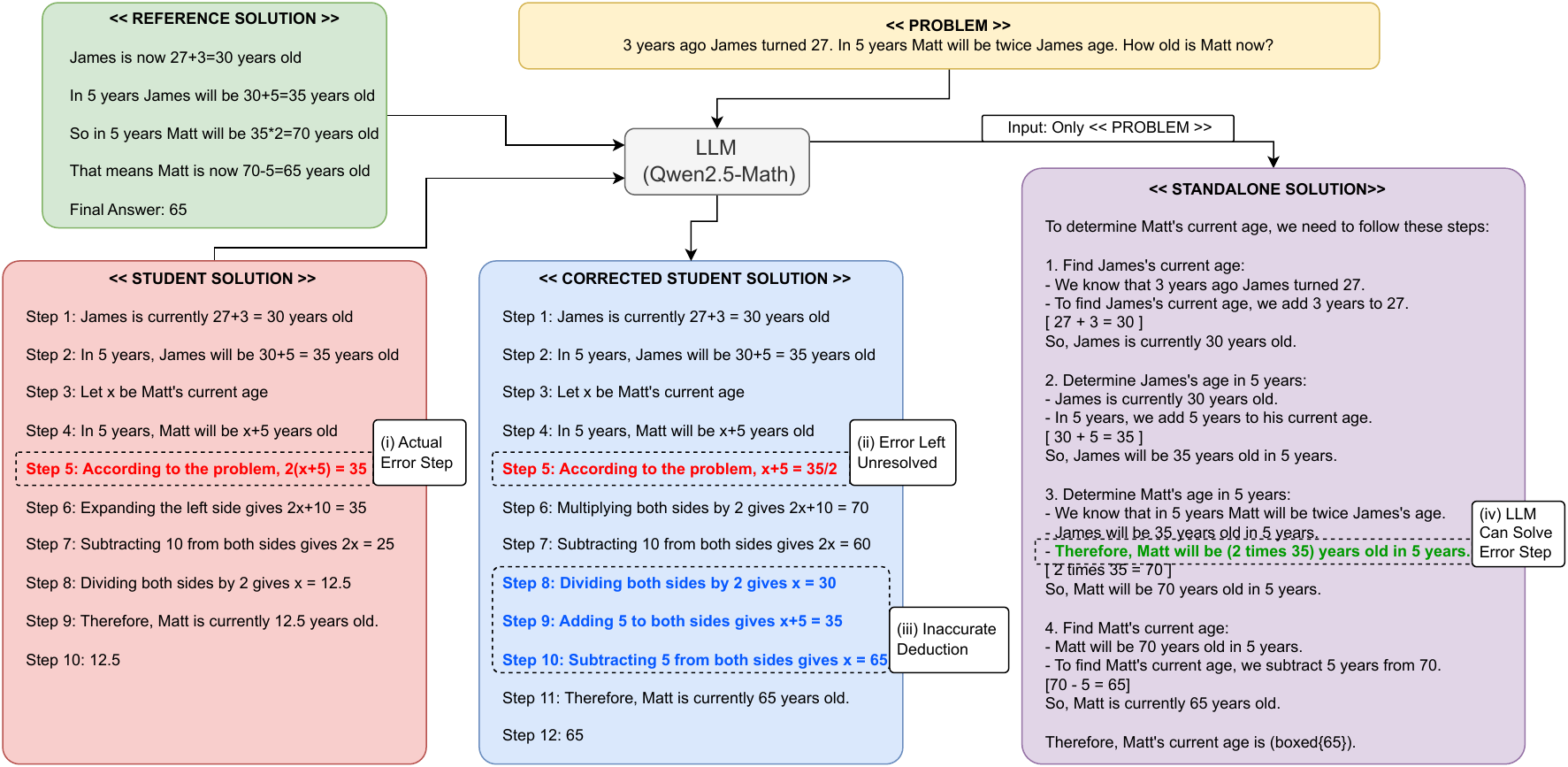}
     \caption{{\tt Qwen2.5-72B-Math} is often unable to rectify the first error step (i) in the student's solution when generating the corrected solution (ii). Instead, additional erroneous deductions (iii) are made later in the solution to make sure that the final answer matches that of the gold (reference) solution. Note that the model can correctly solve the corresponding step in a standalone problem-solving setup (iv).}
     \label{fig:qwen-example}
     \vspace{-6pt}
\end{figure*}

A follow-up qualitative analysis of the predictions of {\tt Qwen2.5-72B-Math} reveals that while generating the corrected solution ({\em w-Cor}), the model often fails to rectify the first error step in student solutions and instead produces inaccurate deductions later in the solution, possibly hallucinating to ensure that the final answer matches that of the gold solution. We present one such example in Figure \ref{fig:qwen-example}. We see that the actual error step (i) remains erroneous in the corrected solution (ii). The model later generates multiple contradictory values of $x$ (iii). This inaccurate correction, in turn, leads to an incorrect error-step prediction. Note that {\tt Qwen2.5-72B-Math} can correctly solve the underlying math problem by itself, including the exact step (iv), which corresponds to the student's first error step.



\subsection{Feature Importance Analysis}
\label{subsec:feature-importance-analysis}

Following prior work on interpretability for black-box models \cite{thakur-etal-2025-mirage,dang2024curiouscasesearchingcorrelation}, we train a Random Forest classifier to predict whether a student error will be correctly localized, using key features related to the problem, solution, and error. We favor a Random Forest model over a Linear Regressor based on their F1-scores as goodness-of-fit proxies (0.996 and 0.572 respectively). The feature set includes linguistic attributes of the math problem (e.g., FKGL, constituency tree depth), the complexity of the gold solution (e.g., counts and types of operations), and descriptors of the student error (e.g., error type and position). We also include a semantic alignment estimate, {\tt Semantic Recall} (§\ref{subsec:app-bertscore-alignment}), measuring how well the reference solution (\(G\) for \textit{w-GS}, \(S'\) for \textit{w-Cor}) aligns with the student’s work up to the first error. See §\ref{subsec:app-feature-importance-analysis-details} for feature set definitions and detailed analysis description.

\begin{figure}[t!]
     \centering
    \includegraphics[width=\columnwidth]{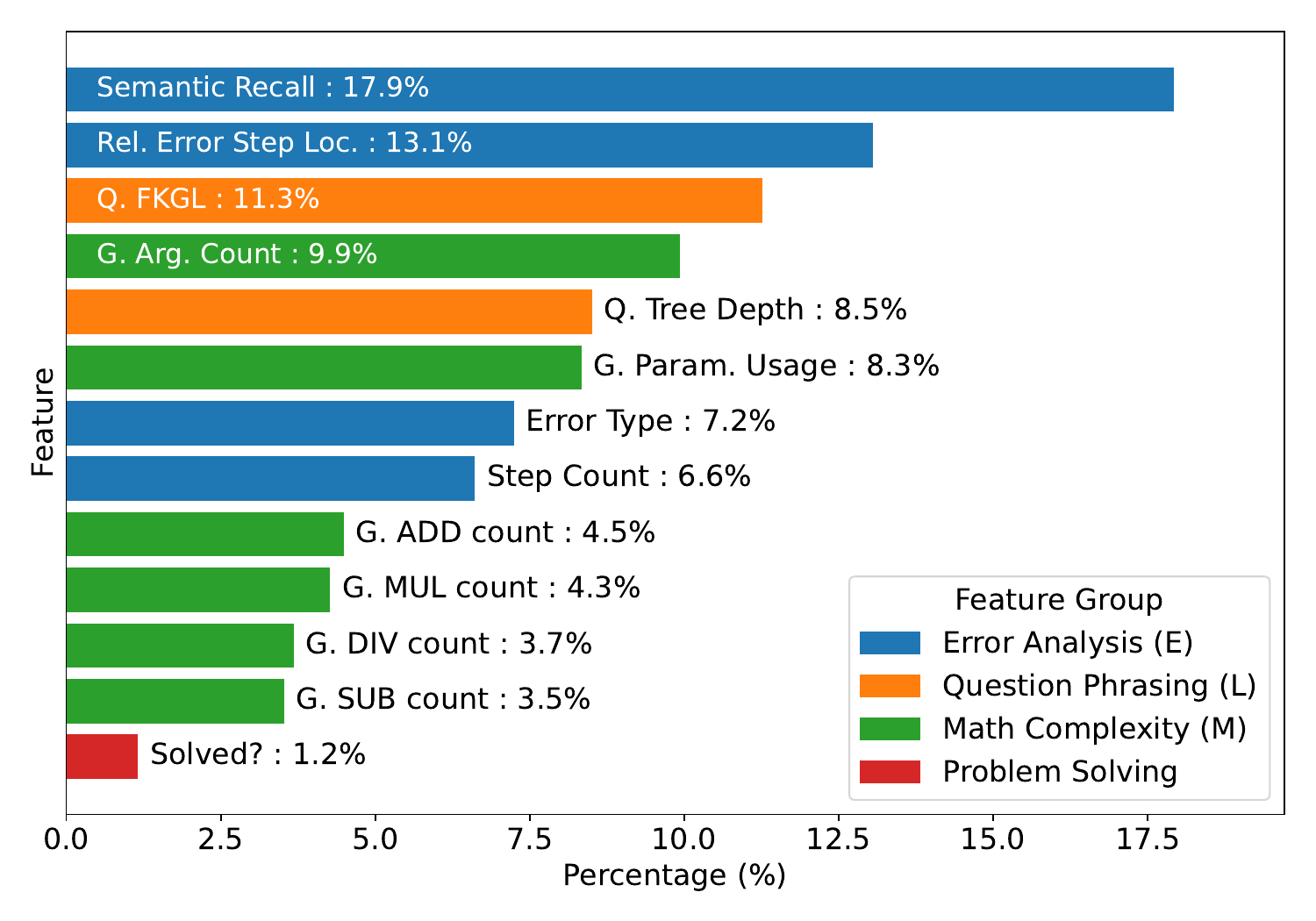}
     \caption{Relative Importance of Features Toward Correct Error Localization}
     \label{fig:feat-imp}
     \vspace{-6mm}
\end{figure}

Figure \ref{fig:feat-imp} shows mean feature importances. The features related to question phrasing (orange) and math complexity (green) are some of the most informative. However, the two most informative features pertain to the error made (blue). {\tt Semantic Recall} is the most important (17.9\%), highlighting the role of alignment in successful error localization. The relative position of the error ({\tt Rel. Error Step Loc.}, 13.1\%) and error type ({\tt Error Type}, 7.2\%) also rank highly.

Interestingly, whether the LLM solved the problem correctly ({\tt Solved?}) has low importance (1.1\%). A chi-squared test for independence confirms that error localization and problem-solving accuracy are weakly correlated (p > 0.01; $\phi$ < 0.2) across models and prompts (see §\ref{subsec:app-chi-square-test} for more details), suggesting that LLMs are not guaranteed to localize errors correctly even when they can solve the underlying problem.

\begin{figure}[t]
     \centering
     \includegraphics[width=\columnwidth]{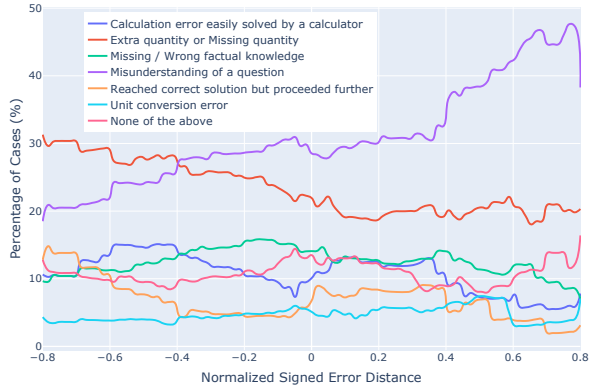}
     \caption{Distribution of ground-truth error types in \texttt{VtG} across models' normalized error step distance}
     \vspace{-0.6cm}
     \label{fig:dist_vs_type}
\end{figure}

\subsection{Error Location vs. Type}
\label{subsec:where-do-llms-go-wrong}
Finally, we examine cases where models' predicted error steps increasingly deviate from the actual steps. We define the normalized error-step distance as the difference between the predicted and actual error steps divided by the total number of steps in the student's solution. In Figure \ref{fig:dist_vs_type}, we plot the distribution of ground-truth error types from \texttt{VtG} against the combined normalized error-step distance across models and prompt types. A distribution shifted to the right indicates that models tend to overshoot the actual error step, while a leftward shift indicates undershooting. We observe that question-independent errors, such as calculation or unit-conversion mistakes, are uniformly distributed regardless of prediction deviation, whereas errors resulting from question misunderstanding are predicted much later than they occur. In contrast, errors involving missing or extra variables tend to be predicted a little before they occur. \textit{This suggests that error-step prediction strategies should also account for the type of error.}

\section{Conclusions}

In this paper, we explored whether incorporating gold solutions enhances LLMs' ability to pinpoint errors in student math solutions from {\tt VtG} and {\tt PRM800K} datasets. Gold solutions improve performance compared to using only the problem and student response, though scores remain low. Replacing the gold solution with an intermediate corrected student solution further boosts performance—especially for smaller models—even though error localization still lags behind overall problem-solving accuracy. Our analysis shows that high problem-solving ability does not guarantee effective error detection, highlighting the need for targeted meta-reasoning improvements. 
Our feature-led analysis also shows that the alignment between the incorrect solution and the reference supports better error localization.
These insights will guide our future work in enhancing LLM performance on error localization and related meta-reasoning tasks.

\section*{Limitations}
This work is subject to several limitations that frame the scope of our findings. First, our experiments are confined to the math domain. While using math word problems provides a controlled setting to explore error localization, it remains unclear whether the observed challenges and benefits would generalize to other domains requiring different reasoning strategies. Second, the study depends on corrections generated by LLMs that are guided by a ground-truth solution. Although these corrected solutions were confirmed to yield the correct final answer, they may still harbor inconsistencies in their intermediate steps. Expert-annotated corrections, which could potentially offer a more reliable reference, were not employed due to the considerable resources required. Third, our evaluation uses a targeted prompting setup designed for comparability across models. Advanced prompting strategies—such as tree-of-thought prompting—have not been explored in this study, leaving open the possibility that alternative approaches might impact error localization performance. Finally, the study is limited to English-language math problems. Given that error localization performance is already challenged in English, it is plausible that the difficulties would be exacerbated in languages with less extensive data representation.

\section*{Ethical Statement}
As the scope of our study is solely to evaluate LLM performance and does not involve private data or manual data creation, we do not foresee any major ethical implications of our work. However, LLMs inherently present risks. These models may generate outputs that, despite being plausible, are factually inaccurate or nonsensical. Such hallucinations can lead to misguided decision-making and the propagation of biases, particularly in high-stakes contexts where accuracy is paramount. In the absence of appropriate safeguards, the broad deployment of LLMs could exacerbate these issues. Thus, it is imperative to develop mechanisms that mitigate the risks of hallucinations to ensure the responsible and effective application of these models.


\bibliography{custom}


\appendix

\section{Dataset Details}
\label{sec:app-data-details}
This section provides further details about the datasets used in our experiments. Tables \ref{tab:stats-for-vtg} and \ref{tab:stats-for-prm800k} provide the distribution statistics of various aspects for \texttt{VtG} and \texttt{PRM800K} respectively, including the number of steps in the student and gold solutions and the location of error steps. Table \ref{tab:vtg-type-perc} lists the error types annotated in {\tt VtG} and their corresponding share of cases in the dataset. There is no associated human labeling for error type in {\tt PRM800K}. Both \texttt{VtG} (CC BY 4.0 License) and \texttt{PRM800K} (MIT License) are publicly accessible datasets.

\begin{table}[h!]
\centering
\resizebox{\columnwidth}{!}{%
\begin{tabular}{@{}lccccc@{}}
\toprule
\textbf{Dimension} & \textbf{Min} & \textbf{Max} & \textbf{Median} & \textbf{$\mu$} & \textbf{$\sigma$} \\ \midrule
Gold Solution Word Length & 6 & 125 & 49 & 50.51 & 20.13 \\ \midrule
Student Solution Word Length & 48 & 109 & 73 & 73.62 & 10.97 \\ \midrule
Gold Solution Step Length & 3 & 5 & 4 & 4.27 & 0.73 \\ \midrule
Student Solution Step Length & 3 & 15 & 5 & 5.92 & 1.84 \\ \midrule
First Error Step Index & 1 & 9 & 3 & 2.77 & 1.43 \\ \bottomrule
\end{tabular}%
}
\caption{Key distributional statistics for \texttt{VtG}}
\label{tab:stats-for-vtg}
\end{table}

\begin{table}[h!]
\centering
\resizebox{\columnwidth}{!}{%
\begin{tabular}{@{}lccccc@{}}
\toprule
\textbf{Dimension} & \textbf{Min} & \textbf{Max} & \textbf{Median} & \textbf{$\mu$} & \textbf{$\sigma$} \\ \midrule
Gold Solution Word Length & 1 & 441 & 72 & 87.75 & 67.82 \\ \midrule
Student Solution Word Length & 6 & 1470 & 209 & 235.86 & 123.57 \\ \midrule
Gold Solution Step Length & 1 & 62 & 3 & 5.72 & 7.02 \\ \midrule
Student Solution Step Length & 2 & 52 & 12 & 13.31 & 6.52 \\ \midrule
First Error Step Index & 1 & 34 & 5 & 5.81 & 4.29 \\ \bottomrule
\end{tabular}%
}
\caption{Key distributional statistics for \texttt{PRM800K}}
\label{tab:stats-for-prm800k}
\end{table}

\begin{table}[h!]
\centering
\resizebox{\columnwidth}{!}{%
\begin{tabular}{@{}lc@{}}
\toprule
\textbf{Error Type} & \multicolumn{1}{l}{\textbf{Percent of Cases (\%)}} \\ \midrule
\textit{Calculation error easily solved by a calculator} & 12.77 \\
\textit{Extra quantity or Missing quantity} & 23.95 \\
\textit{Missing / Wrong factual knowledge} & 13.97 \\
\textit{Misunderstanding of a question} & 28.64 \\
\textit{Reached correct solution but proceeded further} & 6.99 \\
\textit{Unit conversion error} & 4.99 \\
\textit{None of the above} & 8.68 \\ \bottomrule
\end{tabular}%
}
\caption{Annotated error types for student solutions in {\tt VtG} with their respective percentage of each case}
\label{tab:vtg-type-perc}
\end{table}

\section{Querying Details}
\label{sec:app-querying-details}

\subsection{Prompts}
\label{subsec:app-prompts}
\begin{figure}[t!]
     \centering
    \includegraphics[width=\columnwidth]{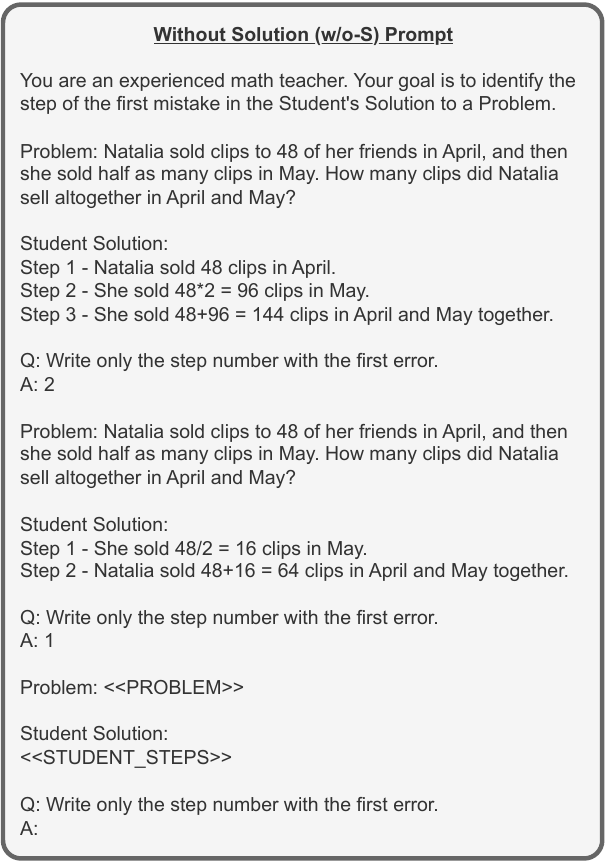}
     \caption{Prompt without solution (\textit{w/o-S})}
     \label{fig:wo-s-prompt}
\end{figure}

\begin{figure}[t!]
     \centering
    \includegraphics[width=\columnwidth]{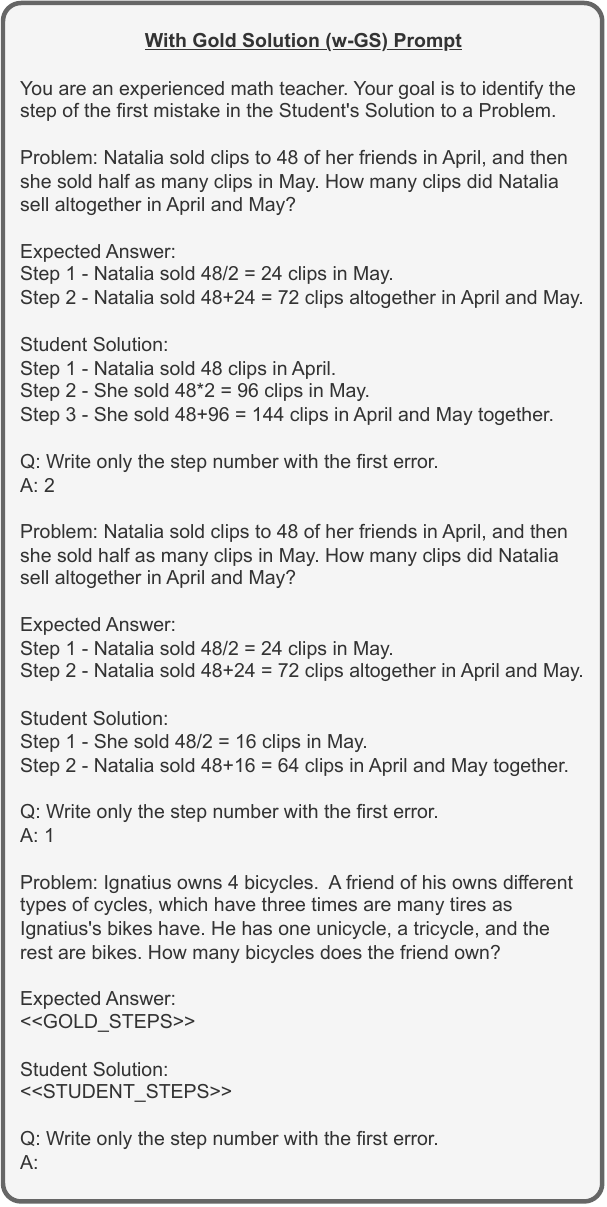}
     \caption{Prompt with gold solution (\textit{w-GS})}
     \label{fig:w-gs-prompt}
\end{figure}

\begin{figure}[t!]
     \centering
    \includegraphics[width=\columnwidth]{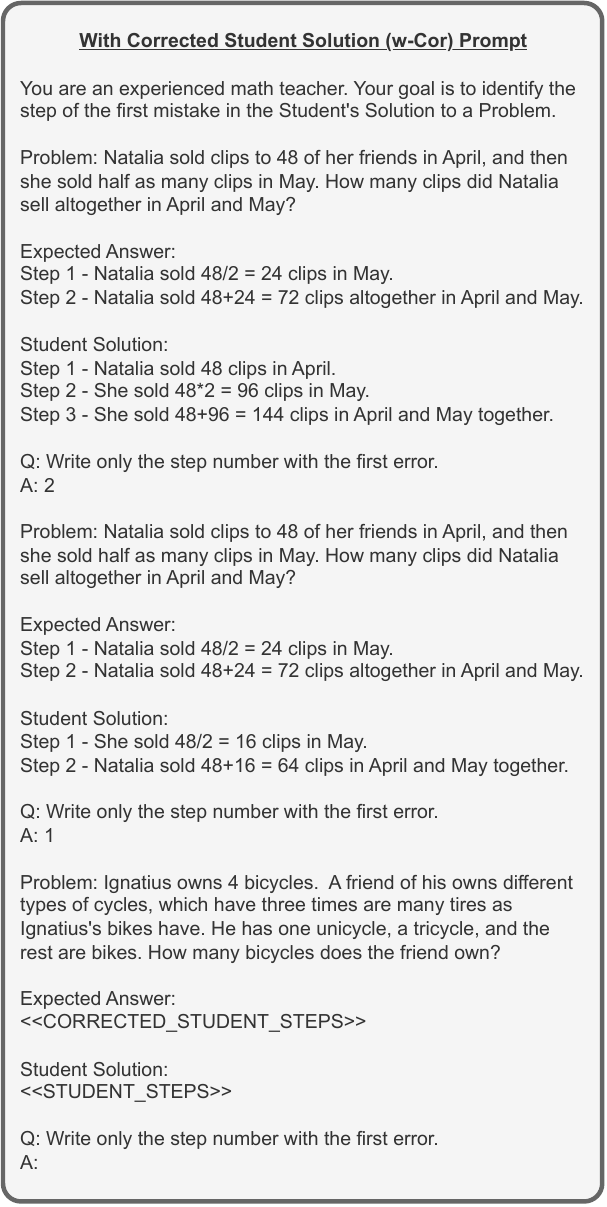}
     \caption{Prompt with corrected student solution (\textit{w-Cor})}
     \label{fig:w-cor-prompt}
\end{figure}

\begin{figure}[t!]
     \centering
    \includegraphics[width=\columnwidth]{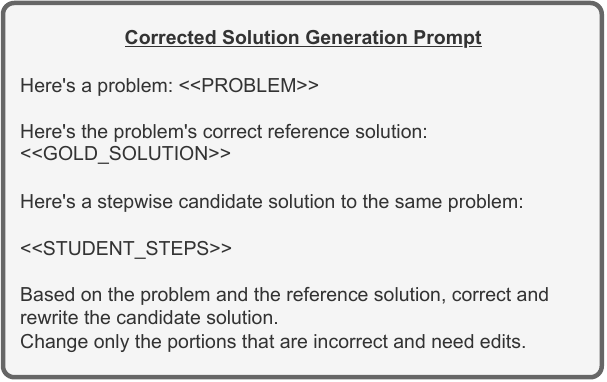}
     \caption{Corrected solution generation prompt}
     \label{fig:correction-generation-prompt}
\end{figure}

This section presents the exact prompts used in our experiments. We begin with the problem-solving prompts used to collect the solutions and final answers to the underlying math questions from {\tt VtG} and {\tt PRM800K} in Figure \ref{fig:problem-solving-prompts}. The initial prompt is used to generate verbose solutions to the questions, followed by a follow-up prompt, where we append the model output with a concluding phrase (i.e., \texttt{Therefore, the final answer is:}) to get the model to specify the final numerical or expression-based answer clearly. We find that this method works best to extract the final answer without additional pattern matching. Next, we show the prompts to predict the exact error-step in the three settings, i.e., without any reference solution (\textit{w/o-S}), with the gold solution (\textit{w-GS}), and with the corrected student solution (\textit{w-Cor}) in Figures \ref{fig:wo-s-prompt}, \ref{fig:w-gs-prompt}, and \ref{fig:w-cor-prompt} respectively. Finally, we show the prompt used to generate the corrected form of the student solutions in Figure \ref{fig:correction-generation-prompt}.

\subsection{Querying Setup}
\label{subsec:app-querying-setup}
This section describes the querying setup used for our experiments. Table \ref{tab:model-versions} shows the exact model versions used. All models were queried with the \texttt{temperature} set to 0, \texttt{top\_p} to 0.95, and \texttt{max\_tokens} to 2048. All \texttt{Llama} models were queried using the Google Cloud (Vertex) API and \texttt{GPT-4o} queries were made using the OpenAI API.

\subsection{Problem-Solving Performance}
\label{subsec:app-problem-solving-performance}
\begin{figure}[t!]
     \centering
    \includegraphics[width=\columnwidth]{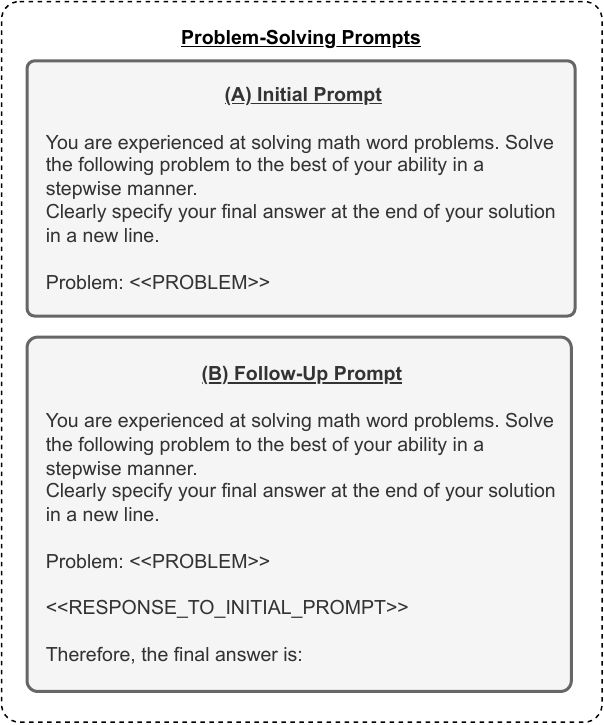}
     \caption{Problem-solving prompts}
     \label{fig:problem-solving-prompts}
\end{figure}

We define LLMs' problem-solving performance as their average accuracy on the math word problems from the test sets of {\tt VtG} and {\tt PRM800K}. Each model is prompted with the math word problem using the prompt templates shown in Figure \ref{fig:problem-solving-prompts} and described in §\ref{subsec:app-prompts}.

\begin{table}[h!]
\centering
\resizebox{0.6\columnwidth}{!}{%
\begin{tabular}{@{}lcc@{}}
\toprule
\textbf{Model} & \textbf{{\tt VtG}} & \textbf{{\tt PRM800K}} \\ \midrule
\texttt{{\tt Llama3-70B}} & 81.04 & 48.15 \\
\texttt{{\tt Llama3.1-70B}} & 88.82 & 62.88 \\
\texttt{{\tt Llama3.1-405B}} & 92.22 & 69.76 \\
\texttt{{\tt GPT-4o}} & 77.45 & 76.22 \\ 
\texttt{{\tt Qwen2.5-72B-Math}} & 83.13 & 87.34 \\ 
\texttt{{\tt LearnLM-1.5-Pro}} & 83.93 & 85.36 \\ \bottomrule
\end{tabular}%
}
\caption{Mean problem-solving accuracy (\%) on the underlying math problems from {\tt VtG} and {\tt PRM800K}}
\label{tab:solving_acc}
\end{table}

\begin{table}[]
\centering
\resizebox{\columnwidth}{!}{%
\begin{tabular}{@{}lccc@{}}
\toprule
\multicolumn{1}{c}{\multirow{2}{*}{\textbf{LLM}}} & \multirow{2}{*}{\textbf{\begin{tabular}[c]{@{}c@{}}Open\\ Source?\end{tabular}}} & \multirow{2}{*}{\textbf{\begin{tabular}[c]{@{}c@{}}Parameter\\ Count\end{tabular}}} & \multirow{2}{*}{\textbf{\begin{tabular}[c]{@{}c@{}}Fine-\\ Tuned?\end{tabular}}} \\
\multicolumn{1}{c}{} &  &  &  \\ \midrule
{\tt LLaMA3-70B} \cite{dubey2024llama} & \cmark{056517} & 70B & \xmark{BF1029} \\ \midrule
{\tt LLaMA3.1-70B} \cite{dubey2024llama} & \cmark{056517} & 70B & \xmark{BF1029} \\ \midrule
{\tt LLaMA3.1-405B} \cite{dubey2024llama} & \cmark{056517} & 405B & \xmark{BF1029} \\ \midrule
{\tt GPT-4o} \cite{openai2024gpt4ocard} & \xmark{BF1029} & -- & \xmark{BF1029} \\ \midrule
\begin{tabular}[c]{@{}l@{}}{\tt Qwen2.5-72B-Math} \cite{yang2024qwen25mathtechnicalreportmathematical}\\\end{tabular} & \cmark{056517} & 72B & \cmark{056517} \\ \midrule
\begin{tabular}[c]{@{}l@{}}{\tt LearnLM-1.5-Pro} \cite{learnlmteam2024learnlmimprovinggeminilearning}\\\end{tabular} & \xmark{BF1029} & -- & \cmark{056517} \\ \bottomrule
\end{tabular}%
}
\caption{The diverse set of LLMs included in this study}
\label{tab:llm-listing}
\end{table}

\begin{table}[]
\centering
\resizebox{\columnwidth}{!}{%
\begin{tabular}{@{}ll@{}}
\toprule
\multicolumn{1}{c}{\textbf{Model}} & \multicolumn{1}{c}{\textbf{Version}} \\ \midrule
\texttt{Llama3-70B} & \texttt{meta-llama/Meta-Llama-3-70B-Instruct} \\
\texttt{Llama3.1-70B} & \texttt{meta-llama/Llama-3.1-70B} \\
\texttt{Llama3.1-405B} & \texttt{meta-llama/Llama-3.1-405B} \\
\texttt{GPT-4o} & \texttt{gpt-4o-2024-08-06} \\ 
\texttt{Qwen2.5-72B-Math} & \texttt{Qwen/Qwen2.5-Math-72B-Instruct} \\ 
\texttt{LearnLM-1.5-Pro} & \texttt{learnlm-1.5-pro-experimental} \\ 
\bottomrule
\end{tabular}%
}
\caption{Model versions for the LLMs used in our experiments}
\label{tab:model-versions}
\end{table}

\section{Additional Analyses \& Details}
\label{sec:app-analysis-details}

\subsection{Alignment with Student Solution}
\label{subsec:app-bertscore-alignment}
We discuss the importance of aligning the ground truth and student solutions for effective comparison and error localization in Section \ref{sec:introduction}. Specifically, we generated intermediate corrected versions of the student solution to serve as ground truth instead of the dataset-provided gold solutions. Ideally, a ground-truth solution should match the student solution up to the first error step, after which divergence is expected. Thus, we measure the semantic overlap between the ground truth and student solutions (truncated before the first error) using {\tt BERTScore} \cite{zhang2020bertscoreevaluatingtextgeneration} recall (see Table \ref{tab:bertscores}). Maximizing recall ensures that the ground truth closely follows the student's approach. We also use this quantity ({\tt Semantic Recall}) as a feature for further analysis in §\ref{subsec:feature-importance-analysis}. \textit{The results show that corrected solutions from all models yield much higher recall than the corresponding gold solutions for both datasets, indicating superior semantic and stylistic alignment with the student solutions.}

\begin{table}[t!]
\centering
\resizebox{0.8\columnwidth}{!}{%
\begin{tabular}{@{}llcc@{}}
\toprule
\textbf{\begin{tabular}[c]{@{}l@{}}Solution\\ Type\end{tabular}} & \textbf{Model} & \texttt{VtG} & \texttt{PRM800K} \\ \midrule
Gold & -- & 89.52 & 85.12 \\ \midrule
\multirow{4}{*}{Corrected} & \texttt{Llama3-70B} & 94.77 & 94.46 \\
 & \texttt{Llama3.1-405B} & 96.04 & 95.34 \\
 & \texttt{Llama3.1-70B} & 96.18 & 95.88 \\
 & \texttt{GPT-4o} & 95.86 & 93.79 \\ 
 & \texttt{Qwen2.5-72B-Math} & 95.03 & 87.66 \\
 & \texttt{LearnLM-1.5-Pro} & 94.98 & 92.98 \\ \bottomrule
\end{tabular}%
}
\caption{{\tt BERTScore} recall between ground-truth solutions (gold or corrected) and student solutions. Recall values for corrected solutions from different LLMs have been recorded separately.}
\label{tab:bertscores}
\end{table}

\subsection{Manual Verification of Generated Corrections}
\label{subsec:app-manual-verification}

We test whether an intermediate generation of a corrected student solution using the gold solution and the student solution serves as a better reference for error localization than the gold solution itself. 

The annotation for each correction involves two questions:

\begin{itemize}
    \item \textbf{Correctness:} Is the LLM-generated correction factually and mathematically sound at each step and does it arrive at the correct answer? (Yes/No)
    \item \textbf{Stylistic Similarity:} Is the LLM-generated correction, stylistically and in approach, similar to the student’s solution up to the first error step? (Yes/No)
\end{itemize}

Table \ref{tab:manual-verification} shows the average percentage values for the two questions for each LLM. The annotation set spans 90 samples (15 samples per LLM), with 30 randomly selected samples (of the 90) to build the agreement subset. We conducted a two-person annotation where both annotators hold at least a masters degree in a STEM field. The first annotator annotates all 90 samples and the second annotator annotates just the agreement set. With this, we estimate that the inter-annotator agreement is $\kappa$ =  0.82 for correctness and 0.85 for stylistic similarity.

\begin{table}[]
\centering
\resizebox{0.8\columnwidth}{!}{%
\begin{tabular}{@{}lcc@{}}
\toprule
\textbf{Model} & \textbf{Correctness} & \textbf{Stylistic Similarity} \\ \midrule
{\tt Llama3-70B} & 93.75 & 91.02 \\ \midrule
{\tt Llama3.1-70B} & 96.67 & 93.33 \\ \midrule
{\tt Llama3.1-405B} & 94.00 & 88.40 \\ \midrule
{\tt GPT-4o} & 93.33 & 89.67 \\ \midrule
{\tt Qwen2.5-72B-Math} & 69.59 & 63.33 \\ \midrule
{\tt LearnLM-1.5-Pro} & 95.71 & 87.43 \\ \bottomrule
\end{tabular}%
}
\caption{Average percentage values of correctness and stylistic similarity based on manual annotation on LLM-generated corrected student solutions}
\label{tab:manual-verification}
\end{table}

\subsection{Chi-Square Test}
\label{subsec:app-chi-square-test}

In §\ref{subsec:feature-importance-analysis}, we present the correlation between LLMs' problem-solving performance and error localization performance. As both variables are binary categorical variables, the appropriate method to determine their relation is to construct a 2$\times$2 contingency table and perform a Chi-Square test to check for a statistically significant association between them \cite{siegel1956nonparametric}. Table \ref{tab:s_and_e_corr} presents the results for the test ($\chi^2$ statistic, p-value, and $\phi$-coefficient) across models, datasets, and prompt types. We interpret the correlation in each case by the corresponding $p$-values and $\phi$-coefficients. A high $p$-value (>0.01) is indicative of poor statistical significance and a low $\phi$-coefficient (<0.2) is indicative of a low effect size (\citet{cohen1988, rea1992} regard coefficient values <0.2 as weak). 

We observe that no setting in Table \ref{tab:s_and_e_corr} yields a strong correlation. \textit{This means that, within an LLM’s set of responses, solving the problem correctly does not strongly predict model's ability to pinpoint an error.} E.g., even though {\tt Qwen2.5-72B-Math} outperforms most other models in problem solving across datasets (see Table \ref{tab:solving_acc}), its error localization performance is the poorest. This further motivates the need for LLMs tuned for better error diagnostic capabilities among other meta-reasoning abilities.

\begin{table}[t!]
\centering
\resizebox{\columnwidth}{!}{%
\begin{tabular}{@{}lcccc|ccc@{}}
\toprule
\multirow{2}{*}{\textbf{Model}} & \multicolumn{1}{c}{\multirow{2}{*}{\textbf{\begin{tabular}[c]{@{}c@{}}Prompt \\ Type\end{tabular}}}} & \multicolumn{3}{c|}{\textbf{{\tt VtG}}} & \multicolumn{3}{c}{\textbf{{\tt PRM800K}}} \\ \cmidrule(l){3-8} 
 & \multicolumn{1}{c}{} & \multicolumn{1}{c}{\textbf{$\chi^{2}$}} & \multicolumn{1}{c}{p-value} & \multicolumn{1}{c|}{\textbf{$\phi$}} & \multicolumn{1}{c}{\textbf{$\chi^{2}$}} & \multicolumn{1}{c}{p-value} & \multicolumn{1}{c}{\textbf{$\phi$}} \\ \midrule
\multirow{3}{*}{\texttt{{\tt Llama3-70B}}} & \textit{w/o-S} & 4.30 & 0.038 & 0.068 & 3.55 & 0.059 & 0.043 \\
 & \textit{w-GS} & 8.00 & 0.005 & 0.092 & 1.34 & 0.247 & 0.027 \\
 & \textit{w-Cor} & 0.00 & 1.000 & 0.002 & 4.30 & 0.038 & 0.047 \\ \midrule
\multirow{3}{*}{\texttt{{\tt Llama3.1-70B}}} & \textit{w/o-S} & 0.49 & 0.482 & -0.025 & 10.09 & 0.001 & 0.071 \\
 & \textit{w-GS} & 0.24 & 0.620 & 0.019 & 16.49 & 0.000 & 0.090 \\
 & \textit{w-Cor} & 0.08 & 0.774 & 0.012 & 29.70 & 0.000 & 0.121 \\ \midrule
\multirow{3}{*}{\texttt{{\tt Llama3.1-405B}}} & \textit{w/o-S} & 0.08 & 0.768 & -0.013 & 3.59 & 0.058 & 0.043 \\
 & \textit{w-GS} & 3.34 & 0.067 & -0.062 & 12.19 & 0.000 & 0.078 \\
 & \textit{w-Cor} & 0.05 & 0.824 & -0.011 & 27.71 & 0.000 & 0.117 \\ \midrule
\multirow{3}{*}{\texttt{{\tt GPT-4o}}} & \textit{w/o-S} & 2.14 & 0.143 & 0.049 & 7.28 & 0.007 & 0.060 \\
 & \textit{w-GS} & 0.94 & 0.332 & 0.033 & 4.51 & 0.034 & 0.048 \\
 & \textit{w-Cor} & 1.77 & 0.183 & 0.045 & 6.92 & 0.009 & 0.059 \\ \midrule
 \multirow{3}{*}{\tt{Qwen2.5-72B-Math}} & \textit{w/o-S} & 17.356 & 0.000 & 0.134 & 3.077 & 0.079 & 0.040 \\
 & \textit{w-GS} & 0.030 & 0.863 & 0.008 & 9.840 & 0.002 & 0.070 \\
 & \textit{w-Cor} & 4.880 & 0.027 & -0.073 & 5.042 & 0.025 & 0.051 \\ \midrule
\multirow{3}{*}{\tt{LearnLM-1.5-Pro}} & \textit{w/o-S} & 13.021 & 0.000 & 0.117 & 3.426 & 0.064 & 0.042 \\
 & \textit{w-GS} & 5.149 & 0.023 & 0.075 & 2.056 & 0.152 & 0.033 \\
 & \textit{w-Cor} & 3.208 & 0.073 & 0.059 & 2.198 & 0.138 & 0.034 \\ \bottomrule
\end{tabular}%
}
\vspace{-0.2cm}
\caption{\small Chi-Square test and $\phi$-coefficient statistics between models' problem-solving performance and error localization performance on {\tt VtG} and {\tt PRM800K} test sets. A high $p$-value (>0.01) or a low $\phi$-coefficient (<0.2) (i.e., weak effect size) are both indicative of a poor correlation \cite{cohen1988, rea1992}.}
\vspace{-0.5cm}
\label{tab:s_and_e_corr}
\end{table}

\subsection{Feature Importance Analysis Details}
\label{subsec:app-feature-importance-analysis-details}

We fit a Random Forest classifier to predict whether an LLM will be able to correctly predict the first error step in a given student solution using key features to determine their relative importance in determining the LLM's performance. In this section, we describe the feature set that we used and the process of fitting the model and extracting the feature importance scores. 

\paragraph{Feature Set}
We use a feature set capturing the phrasing of the math problem (L), the mathematical complexity of the underlying gold solution (M), and details about the error made by the student (E). We borrow the L and M type features and their exact extraction implementation from \citet{srivatsa-kochmar-2024-makes}. The feature set is as follows: 

\begin{itemize}
    \item {\tt Q. Word Length (L)}: The number of space-separated words in the question text.
    \item {\tt Q. Arg. Count (L)}: Number of distinct numerical quantities in the question text. E.g., ``\textbf{20} boxes'' or ``\textbf{1.5} hours later''.
    \item {\tt Q. FKGL (L)}: The FKGL readability grade \cite{kincaid1975derivation} of the question text.
    \item {\tt Q. Tree Depth (L)}: The average depth of the constituency tree for the sentences in the question text.
    \item {\tt Q. NP Count (L)}: The number of unique noun phrases in the question text.
    \item {\tt G. Arg. Count (M)}: The number of distinct numerical quantities in the gold solution. These may include the arguments imported from the question text and intermediate arguments calculated in the solution steps.
    \item {\tt G. ADD/ SUB/ MUL/ DIV Count (M)}: The number of instances of each of the arithmetic operators used in the gold solution.
    \item {\tt G. Op. Unique Count (M)}: The number of unique arithmetic operators used in the gold solution.
    \item {\tt G. Op. Diversity (M)}: Ratio of {\tt G. Op. Unique Count} and {\tt G. ADD/ SUB/ MUL/ DIV Count}.
    \item {\tt G. Param. Usage (M)}: Ratio of {\tt G. Arg. Count} and {\tt Q. Arg. Count}. This serves as a measure of the proportion of input arguments that are actually relevant to solving the problem. A lower ratio means a greater number of distractors.
    \item {\tt G. World Knowledge (M)}: The number of arguments in the gold solution that are neither input arguments from the question text nor intermediate variables. Such arguments are mainly real world quantities required to solve the problem but not explicitly provided by the question.
    \item {\tt Step Count (E)}: The total number of steps in the incorrect student solution.
    \item {\tt Rel. Error Step Loc. (E)}: The relative position of the first error step in the incorrect student solution. This is defined as the ratio of the first error's step index and the total number of steps in the student solution.
    \item {\tt Error Type (E)}: One of the 7 error types as shown in Table \ref{tab:vtg-type-perc}.
    \item {\tt Semantic Recall (E)}: An estimate of semantic alignment between the steps of the student solution and the reference solution (gold solution for \textit{w-GS} and corrected student solution for \textit{w-Cor}) up to the first error step in the student solution. This is defined as the BERTScore \cite{zhang2020bertscoreevaluatingtextgeneration} recall between the two solutions for the solution texts before the first erroneous step in the student solution. See more details in \ref{subsec:app-bertscore-alignment}.
\end{itemize}

\paragraph{Model Fitting}
We use the Random Forest (RF) implementation from \href{https://scikit-learn.org/stable/modules/generated/sklearn.ensemble.RandomForestClassifier}{{\tt Scikit-Learn}}. Before training the RF model, we prune the feature data to only retain features with an absolute Spearman correlation value <0.4. This removes redundant features, which would otherwise make interpreting feature importance scores difficult. The RF model is trained with 200 estimators and each model and prompt setting is trained 10 times with varying initialization seeds. 

\paragraph{Feature Importance Calculation}
Trained RF models return the normalized (sum = 1.0) Gini importance values for each input feature. The overall importance value ($\Lambda_{i}$ \%) for a feature $i$ across RF models is aggregated as weighted mean of each model's feature importance for feature $i$ ($\lambda_{ij}$), by the corresponding goodness of fit, i.e., accuracy ($a_j$) (see Eq. \ref{eq:weighted_mean}).

\begin{equation}
 \Lambda _{i} = 100 \times \frac{\sum_{j} a_j \cdot \lambda_{ij}}{\sum_{j} a_j}
\label{eq:weighted_mean}
\end{equation}

\subsection{How far off are LLMs?}
\label{subsec:how-far-off-are-we}
We aim to assess how close the predicted error step is to the true error step when the prediction is incorrect. To do so, we compute the percentage of incorrect predictions that fall within ±1 and ±2 steps of the actual first error step, as detailed in Table \ref{tab:error-dist} and further distributions in §\ref{subsec:app-error-step-distance-distribution}. Our analysis reveals that for \texttt{VtG} (median step count: 5), between 45\% and 60\% of the incorrect predictions are within ±1 step, whereas for {\tt PRM800K} (median step count: 12), approximately 25\% fall within ±1 step and nearly 50\% within ±2 steps. Additionally, among the three prompt settings, \textit{w-Cor} most consistently achieves the highest number of predictions within both windows and performance across models is similar, with \texttt{Llama3-70B} matching or surpassing both \texttt{GPT-4o} and \texttt{Llama3.1-405B}. \textit{These results suggest that while models often miss the exact first error step, their predictions remain close, motivating the development of fine-grained policies to precisely pinpoint error steps in future work.}

\begin{table}[]
\centering
\resizebox{0.85\columnwidth}{!}{%
\begin{tabular}{@{}lccc|cc@{}}
\toprule
\multirow{2}{*}{\textbf{Model}} & \multirow{2}{*}{\textbf{\begin{tabular}[c]{@{}c@{}}Prompt \\ Type\end{tabular}}} & \multicolumn{2}{c|}{\textbf{{\tt VtG}}} & \multicolumn{2}{c}{\textbf{{\tt PRM800K}}} \\ \cmidrule(l){3-6} 
 &  & ±1 & \multicolumn{1}{c|}{±2} & ±1 & ±2 \\ \midrule
Random & \textit{--} & 31.89 & \multicolumn{1}{c|}{55.5} & 17.53 & 32.42 \\ \midrule
\multirow{3}{*}{\tt{Llama3-70B}} & \textit{w-GS} & 56.47 & \multicolumn{1}{c|}{82.64} & 28.17 & 46.78 \\
 & \textit{w-GS} & 56.47 & \multicolumn{1}{c|}{82.64} & 28.17 & 46.78 \\
 & \textit{w-Cor} & \textbf{59.20} & \multicolumn{1}{c|}{\textbf{84.48}} & \textbf{31.25} & \textbf{49.36} \\ \midrule
\multirow{3}{*}{\tt{Llama3.1-70B}} & \textit{w/o-S} & 57.28 & \multicolumn{1}{c|}{\textbf{82.82}} & 24.52 & 41.91 \\
 & \textit{w-GS} & 53.70 & \multicolumn{1}{c|}{78.84} & \textbf{27.5} & \textbf{45.38} \\
 & \textit{w-Cor} & \textbf{58.27} & \multicolumn{1}{c|}{81.30} & 25.97 & 41.70 \\ \midrule
\multirow{3}{*}{\tt{Llama3.1-405B}} & \textit{w/o-S} & 47.61 & \multicolumn{1}{c|}{77.68} & 27.16 & 45.49 \\
 & \textit{w-GS} & 54.86 & \multicolumn{1}{c|}{82.00} & 26.72 & \textbf{48.03} \\
 & \textit{w-Cor} & \textbf{56.01} & \multicolumn{1}{c|}{\textbf{83.58}} & \textbf{30.79} & 47.85 \\ \midrule
\multirow{3}{*}{\tt{GPT-4o}} & \textit{w/o-S} & 56.21 & \multicolumn{1}{c|}{83.16} & 25.36 & 43.31 \\
 & \textit{w-GS} & 53.12 & \multicolumn{1}{c|}{78.30} & \textbf{28.92} & \textbf{49.18} \\
 & \textit{w-Cor} & \textbf{60.86} & \multicolumn{1}{c|}{\textbf{84.57}} & 23.48 & 37.52 \\ \midrule
\multirow{3}{*}{\tt{Qwen2.5-72B-Math}} & \multicolumn{1}{l}{w/o-S} & \multicolumn{1}{l}{\textbf{48.36}} & \multicolumn{1}{l|}{73.24} & \multicolumn{1}{l}{37.78} & \multicolumn{1}{l}{47.54} \\
 & \multicolumn{1}{l}{w-GS} & \multicolumn{1}{l}{46.35} & \multicolumn{1}{l|}{\textbf{74.92}} & \multicolumn{1}{l}{\textbf{48.97}} & \multicolumn{1}{l}{\textbf{58.51}} \\
 & \multicolumn{1}{l}{w-Cor} & \multicolumn{1}{l}{32.68} & \multicolumn{1}{l|}{65.08} & \multicolumn{1}{l}{43.69} & \multicolumn{1}{l}{56.37} \\ \midrule
\multirow{3}{*}{\tt{LearnLM-1.5-Pro}} & \multicolumn{1}{l}{w/o-S} & \multicolumn{1}{l}{54.36} & \multicolumn{1}{l|}{80.77} & \multicolumn{1}{l}{61.70} & \multicolumn{1}{l}{72.22} \\
 & \multicolumn{1}{l}{w-GS} & \multicolumn{1}{l}{\textbf{58.14}} & \multicolumn{1}{l|}{\textbf{83.72}} & \multicolumn{1}{l}{66.11} & \multicolumn{1}{l}{75.15} \\
 & \multicolumn{1}{l}{w-Cor} & \multicolumn{1}{l}{57.26} & \multicolumn{1}{l|}{83.24} & \multicolumn{1}{l}{\textbf{66.94}} & \multicolumn{1}{l}{\textbf{75.66}} \\ \bottomrule
\end{tabular}%
}
\vspace{-0.1cm}
\caption{\small Percentage of incorrect first error-step predictions where the prediction lies within ±1 and ±2 steps of the actual first error step. \textit{Bold} values denote the greatest percentage value among the three prompt settings for a given model and dataset.}
\vspace{-0.7cm}
\label{tab:error-dist}
\end{table}

\subsection{Error-Step Distance Distribution}
\label{subsec:app-error-step-distance-distribution}

\begin{figure*}[t!]
     \centering
    \includegraphics[width=1\textwidth]{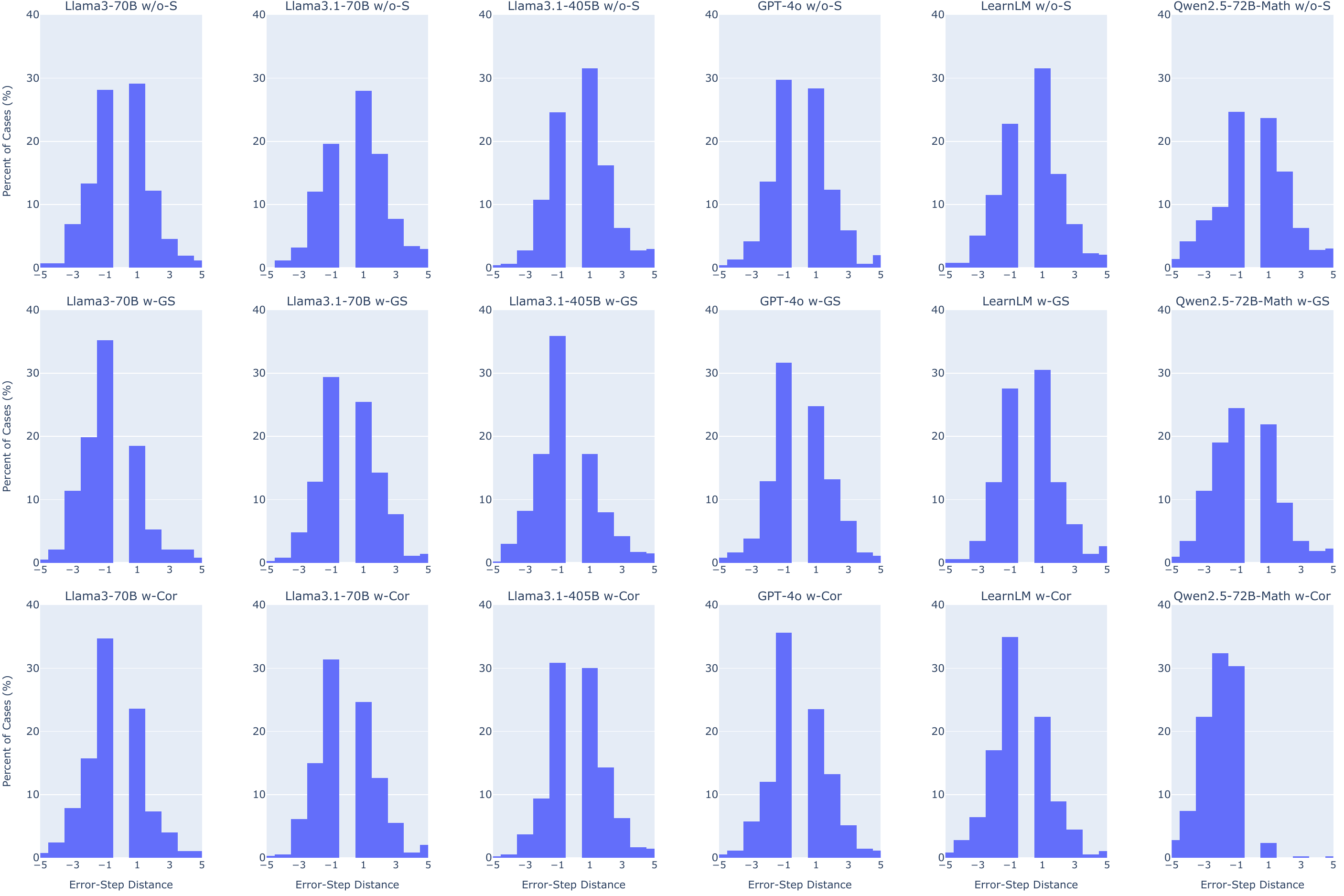}
     \caption{Error-step distance distributions for {\tt VtG}}
     \label{fig:error-step-dist-distribution_eth-vtg}
\end{figure*}

\begin{figure*}[]
     \centering
    \includegraphics[width=1\textwidth]{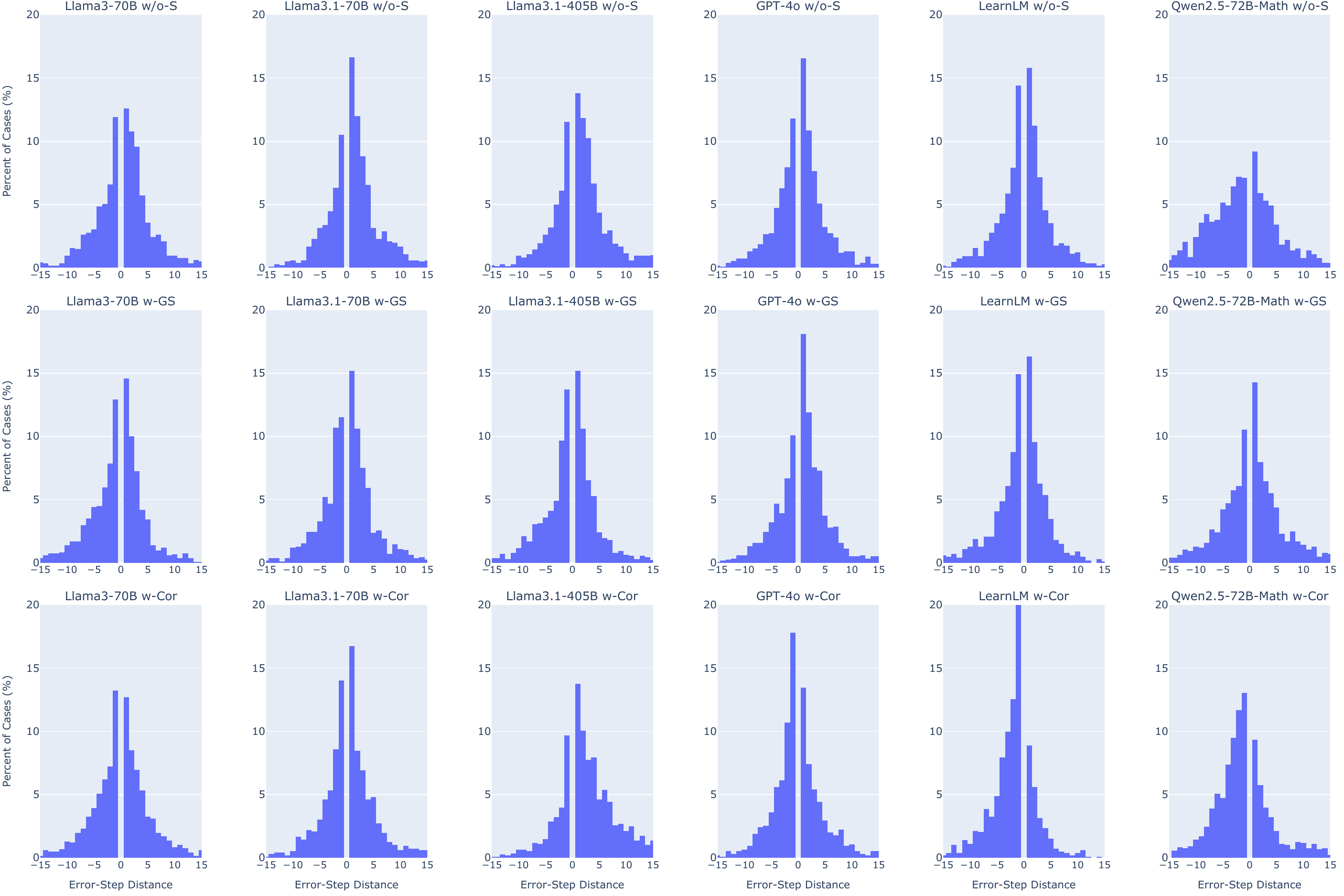}
     \caption{Error-step distance distributions for {\tt PRM800K}}
     \label{fig:error-step-dist-distribution_prm800k}
\end{figure*}

In §\ref{subsec:how-far-off-are-we}, we report the proportion of incorrectly predicted error steps that lie within ±1 and ±2 steps of the actual error step. In Figures \ref{fig:error-step-dist-distribution_eth-vtg} and \ref{fig:error-step-dist-distribution_prm800k}, we present a more detailed distribution of incorrectly predicted error steps by their relative distance from the actual error step for both datasets.




\end{document}